\documentclass{article} 
\usepackage[margin=1in]{geometry}
\usepackage{times}

\usepackage{amsmath,amsfonts,bm}

\def\eqref#1{equation~\ref{#1}}

\def\1{\bm{1}}

\DeclareMathAlphabet{\mathsfit}{\encodingdefault}{\sfdefault}{m}{sl}
\SetMathAlphabet{\mathsfit}{bold}{\encodingdefault}{\sfdefault}{bx}{n}

\usepackage{hyperref}
\usepackage{url}
\usepackage{graphicx}
\usepackage{booktabs}
\usepackage{algorithm}
\usepackage{algorithmic}
\usepackage{float}
\usepackage{needspace}
\usepackage{wrapfig}

\usepackage{natbib}

\title{TRAC: Trajectory-aware Reuse and Adaptive Correction for Efficient Autoregressive Video Generation}

\author{
  Jiaxing Song$^{1}$, Weiqi Yan$^{2}$, You Huang$^{2}$, Mingte Qiu$^{1}$, Huazhong Liu$^{1,3}$, Xiaofeng Zhu$^{1}$, Yunshan Zhong$^{1,*}$ \\
  \vspace{1.5mm} \\
  $^1$School of Computer Science and Technology, Hainan University \\
  $^2$MAC Lab, Department of Artificial Intelligence, School of Informatics, Xiamen University \\
  $^3$Zhejiang Normal University, Jinhua, China \\
  \vspace{1.5mm} \\
  $^*$Corresponding author: \texttt{yszhong01@gmail.com}
}

\begin{document}

\maketitle

\begin{abstract}

Autoregressive (AR) video generation enables scalable long-video synthesis by sequentially generating fixed-length chunks, but incurs substantial computational cost from repeated denoising and attention over growing historical contexts. 
Recent cache reuse methods exploit temporal redundancy to reduce this cost, yet are mainly designed for single-trajectory generation, making them less suitable for sequentially coupled AR trajectories where approximation errors can accumulate and propagate across chunks.
To address this challenge, we propose \textbf{TRAC}, a \textbf{T}rajectory-aware \textbf{R}euse and \textbf{A}daptive \textbf{C}orrection framework for efficient and robust AR video generation. TRAC integrates three complementary components: 
(1) \textbf{Robust Cumulative Scheduling (RCS)}, which selects cache reuse schedules by accounting for cumulative rollout errors and their variation across chunks and prompts; 
(2) \textbf{Autoregressive Trajectory-aware Guidance Scheduling (ATGS)}, which coordinates classifier-free guidance refreshes along the global AR trajectory;
and (3) \textbf{Spectral Structure Correction (SSC)}, which adaptively restores low-frequency structural information to mitigate long-term error accumulation.
By combining these innovations, TRAC substantially improves the efficiency-quality trade-off of AR video generation. For example, on VBench with SkyReels-V2, it achieves a $6.03\times$ speedup over the vanilla model with only $\sim0.6\%$ VBench score degradation, while consistently outperforming existing AR acceleration methods in both latency and generation quality. The code is publicly available at \url{https://github.com/yolosjxhhhhhh/trac}.

\end{abstract}

\section{Introduction}

Autoregressive (AR) video generation has recently emerged as a promising paradigm for synthesizing long videos~\citep{yin2025slow,huang2026self}. Instead of denoising an entire video jointly, AR video models divide a long video into fixed-length chunks and generate them sequentially. For each chunk, the model performs a denoising process conditioned on the previously generated chunks, which are retained as historical context, making long-video generation substantially more scalable~\citep{chen2025skyreels,teng2025magi,gao2024ca2}.

However, this scalability comes at a considerable computational cost~\citep{ma2026flow,ji2026forcing,lu2026fademem}.
As generation proceeds, each new chunk must repeatedly attend to the growing historical context and undergo a heavy denoising trajectory. Moreover, classifier-free guidance (CFG) requires both conditional and unconditional predictions at each guided denoising step~\citep{ho2022classifier}, further increasing the inference cost.
Consequently, long-video generation remains far from practical real-time deployment~\citep{teng2025magi,ma2026flow}.
%

To reduce this cost, a growing body of work accelerates video generation by selectively executing the full model at a subset of denoising steps and reusing cached information at the remaining steps~\citep{guo2026efficient,ji2026forcing,shen2026headcast,lv2026light,yi2025deep,lu2026fademem,ma2026flow,xu2026motion}.
Among them, training-free cache reuse-based acceleration methods are particularly attractive because they introduce no additional training or modification to the base generator~\citep{ma2024deepcache,ma2024learning,liu2411timestep,zhao2025real,kahatapitiya2025adaptive,lyu2025fastercache,liu2026d2cache,zhou2025less,xi2025sparse,castillo2025adaptive,lee2026tortoise,gao2026meancache}.
For example, MeanCache~\citep{gao2026meancache} approximates intermediate velocities using cached Jacobian-vector products (JVPs) and determines a cache schedule through offline calibration.
CFG reuse methods~\citep{castillo2025adaptive,lyu2025fastercache,lee2026tortoise} exploit the similarity between conditional and unconditional predictions by refreshing the guidance residual only at selected steps.
These methods demonstrate that substantial redundancy exists throughout the generation process and can be effectively exploited for efficient inference.

Despite this progress, directly applying existing cache reuse methods to AR video generation remains challenging.
Most existing methods are developed for bidirectional video models, where the video is generated through a single denoising trajectory and reuse errors are primarily confined to that trajectory~\citep{ma2026flow,xu2026motion}.
In contrast, AR generation consists of sequentially connected chunk-level trajectories, where the approximation error in an earlier chunk can propagate through the autoregressive context and affect subsequent chunks. In this paper, considering this distinction, we identify three complementary challenges.
\textbf{First, cache scheduling should account for cumulative and cross-sample errors rather than local similarity alone.}
Existing schedulers such as MeanCache~\citep{gao2026meancache} mainly evaluate local reuse errors, which may overlook their accumulation along the generation path and their variation across chunks and prompts.
We therefore introduce robust cumulative scheduling (RCS), which selects cache schedules by jointly considering cumulative rollout error and cross-chunk/prompt risk.
\textbf{Second, CFG reuse should account for the sequential structure of the global AR trajectory.}
Existing CFG reuse methods~\citep{lyu2025fastercache,lee2026tortoise} are mainly designed for bidirectional models with a single denoising trajectory.
In AR generation, however, multiple chunk-level trajectories are sequentially coupled through autoregressive context, making CFG reuse decisions dependent on the complete execution path.
We therefore propose autoregressive trajectory-aware guidance scheduling (ATGS), which calibrates CFG refreshes over the global AR trajectory.
\textbf{Third, cache reuse can induce long-term low-frequency structural loss.}
Repeated approximation gradually accumulates errors in low-frequency components that encode global appearance and layout~\citep{li2026freqforcing}.
We therefore introduce spectral structure correction (SSC), which uses the first chunk as a stable structural reference and adaptively injects low-frequency information according to local motion, preserving global coherence while retaining dynamic details.

As shown in Fig.~\ref{fig:main}, RCS, ATGS, and SSC form a complementary framework, termed trajectory-aware reuse and adaptive correction (TRAC), for efficient and robust AR video generation.
By jointly reducing redundant computation and controlling error accumulation, TRAC substantially accelerates long-video generation while maintaining high visual fidelity and temporal consistency. 
On VBench with SkyReels-V2, TRAC achieves a $6.03\times$ speedup over the vanilla model with only $\sim0.6\%$ quality degradation (82.44 vs.~82.97). Compared to FlowCache-slow~\citep{ma2026flow}, TRAC delivers a $0.36$ higher VBench score while providing substantially better throughput ($6.03\times$ vs.~$4.97\times$). Against MotionCache~\citep{xu2026motion}, TRAC improves the score by $0.64$ while accelerating generation by nearly $29\%$ ($6.03\times$ vs.~$4.68\times$).


\section{Related Work}

\paragraph{Autoregressive Video Generation}
 
Autoregressive video generation has become a dominant paradigm for extending video duration via producing content sequentially in temporally ordered video chunks~\citep{villegas2022phenaki,kondratyuk2023videopoet,henschel2025streamingt2v,weng2024art,gao2024vid}. 
Recent advances in long-video diffusion, including Diffusion Forcing, CausVid, Ca2-VDM, SkyReels-V2, MAGI-1, LongLive, Self Forcing, and FramePack, further underscore the viability of autoregressive and diffusion-forcing approaches for high-fidelity long-video synthesis~\citep{chen2024diffusion,yin2025slow,gao2024ca2,chen2025skyreels,teng2025magi,yang2025longlive,huang2026self,zhang2026frame}.
In these models, the generation process interleaves chunk-local denoising trajectories, where later steps rely on previously generated content as historical context. Effective acceleration techniques for such a setting must therefore address both local denoising dynamics and the cumulative error that propagates across consecutive chunks~\citep{guo2026efficient,ji2026forcing,shen2026headcast,lv2026light,yi2025deep,luo2026future,lu2026fademem,ma2026flow,xu2026motion}.


\paragraph{Diffusion Caching and Denoising Scheduling}

Training‑free cache acceleration methods reduce inference cost by reusing slowly‑varying intermediate features across neighboring denoising steps~\citep{zhao2025real,lyu2025fastercache}. For example, DeepCache reuses U-Net features \citep{ma2024deepcache}, Learning-to-Cache learns a timestep-dependent layer caching policy \citep{ma2024learning}. TeaCache estimates model-output changes from timestep-embedding-modulated inputs \citep{liu2411timestep}. MeanCache constructs interval average velocities from cached JVPs, builds an offline trajectory map, and determines a fixed cache schedule via graph search \citep{gao2026meancache}. The recent FlowCache generalizes uniform timestep caching to chunk‑wise policies \citep{ma2026flow}, with a specific focus on AR video models.
Moreover, to alleviate the computational cost of Classifier‑free guidance~\citep{ho2022classifier}, several approaches exploit temporal redundancy in the guidance residual to reduce this overhead \citep{castillo2025adaptive,lyu2025fastercache,lee2026tortoise}.

\paragraph{Frequency-domain Stabilization for Video Generation}

Extensive methods have also been proposed to address temporal consistency and structural drift in long‑video generation. For instance, FreeNoise reschedules noise for extended video diffusion \citep{qiu2024freenoise}. FreeInit refines initial noise to close the initialization gap~\citep{wu2024freeinit}. GLC‑Diffusion integrates global and local denoising to balance long‑range coherence and fine‑grained detail \citep{ma2026tuning}.
Another thread tackles degradation from a frequency‑domain perspective. For instance, FreeLong~\citep{lu2024freelong} extends video length by spectrally blending global and local temporal attention features, while FreeLong++\citep{lu2025freelong++} improves consistency across multiple frequency bands through a multi-band spectral fusion approach. Most relevant to our work, FreqForcing~\citep{li2026freqforcing} attributes drift in autoregressive long‑video generation to low‑frequency spectral energy, and proposes Spectral Self‑Anchoring (SSA) to address it. SSA combines a local branch for motion/details, an anchor branch for low‑frequency structure from early frames, and frequency‑domain fusion.


\section{Method}
\label{sec:method}

\begin{figure*}[t]
    \centering
    \includegraphics[width=0.85\textwidth]{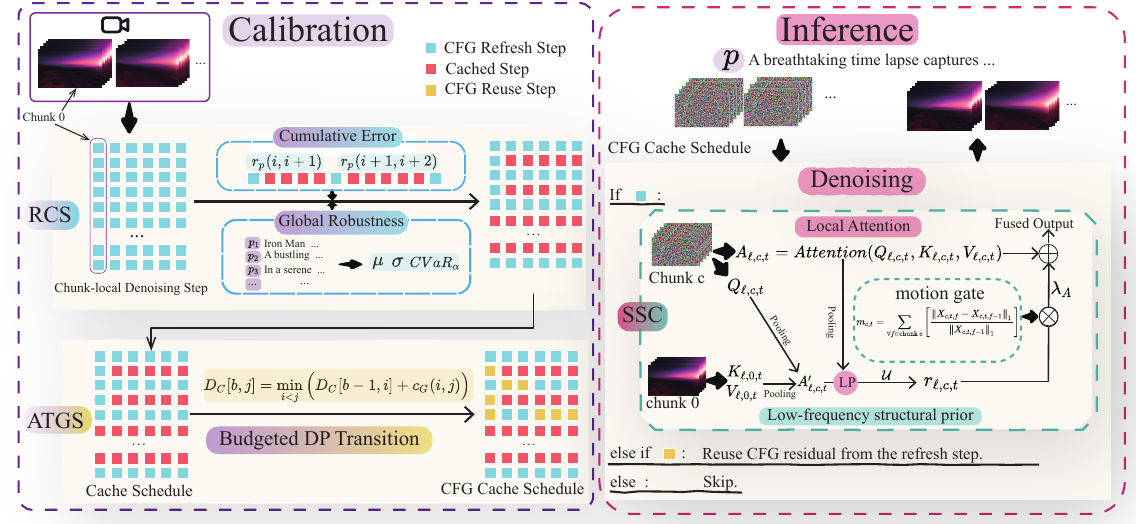}
    \caption{\textbf{Overview of the proposed trajectory-aware reuse and adaptive correction (TRAC).} 
    TRAC comprises three complementary mechanisms for AR video generation, including robust cumulative scheduling (RCS), autoregressive trajectory-aware guidance scheduling (ATGS), and spectral structure correction (SSC).
    %
    }
    \label{fig:main}
\end{figure*}

\subsection{Preliminaries and notation}

\textbf{Autoregressive video model.} Autoregressive (AR) video models generate long videos by decomposing them into multiple fixed-length chunks and generating these chunks sequentially. Let $c\in\{0,\ldots,C-1\}$ denote the chunk index and $t\in\{0,\ldots,T-1\}$ denote the chunk-local denoising step. We denote the latent state of chunk $c$ at step $t$ as $X_{c,t}$ and the historical context from previously generated chunks as $H_c$. A model forward pass is formulated as:
\begin{equation}
v_{c,t}=F_\theta(X_{c,t}, t, \mathrm{cond}, H_c),
\end{equation}
where $\mathrm{cond}$ represents the text, image, or video conditioning.

\textbf{Cache Reuse.}
Cache reuse~\citep{ma2024deepcache,ma2024learning,liu2411timestep,zhao2025real,ma2026flow,gao2026meancache,xu2026motion} accelerates inference by selectively executing full model forward passes at a subset of denoising steps and reusing cached information at the remaining steps. 
Specifically, a cache schedule $\mathcal{S}=\{s_i\}_{i=0}^{L-1}$ specifies the denoising steps where full model forward passes are executed. Two adjacent selected steps $(s_i,s_{i+1})$ define a reuse span, within which intermediate steps reuse cached information. Recently, MeanCache~\citep{gao2026meancache} approximates intermediate velocities via cached Jacobian-vector products (JVPs), which are averaged over a subset of prompts to enable efficient video synthesis.

\textbf{Classifier-free guidance.} Classifier-free guidance (CFG)~\citep{ho2022classifier} combines conditional and unconditional predictions to strengthen conditioning effects. For chunk $c$ at denoising step $t$, the conditional and unconditional predictions are denoted as $\epsilon^{cond}_{c,t}$ and $\epsilon^{uncond}_{c,t}$, respectively. 
The guidance residual is defined as $\Delta_{c,t}=\epsilon^{cond}_{c,t}-\epsilon^{uncond}_{c,t}$, and the guided prediction is computed as $\hat{\epsilon}_{c,t}
=
\epsilon^{cond}_{c,t}
+
(w-1)\Delta_{c,t}$, where $w$ denotes the CFG scale.
To reduce overhead, CFG reuse methods~\citep{castillo2025adaptive,lyu2025fastercache,lee2026tortoise} refresh $\Delta_{c,t}$ only at a subset $\mathcal{S}^C$ of denoising steps, and reuse the most recent cached residual at all other steps:
\begin{equation}
\Delta_{c,t}=
\begin{cases}
\epsilon^{cond}_{c,t}-\epsilon^{uncond}_{c,t}, & t\in \mathcal{S}^C,\\
\Delta_{c,s}, & t\notin \mathcal{S}^C,~s=\max\{j\in \mathcal{S}^C:j<t\},
\end{cases}
\end{equation}
where $\mathcal{S}^C$ denotes the set of steps that require a full forward pass of the model.

\subsection{Trajectory-aware reuse and adaptive correction}

As shown in Fig.~\ref{fig:main}, we propose trajectory-aware reuse and adaptive correction (TRAC), including robust cumulative scheduling (RCS), autoregressive trajectory-aware guidance scheduling (ATGS), and spectral structure correction (SSC), for efficient and robust AR video generation. The overall algorithm is presented in Appendix~\ref{app:Unified algorithm}.

\subsubsection{Robust Cumulative Scheduling}

Despite its progress, MeanCache~\citep{gao2026meancache} determines the cache schedule merely based on the average of local reuse errors over all chunks and prompts, while neglecting cross-chunks/prompts robustness and the cumulative impact of reuse errors along the trajectory. To address these issues, we propose robust cumulative scheduling (RCS), which formulates cache schedule selection as a risk-aware optimization problem that jointly accounts for cross-chunks/prompts robustness and cumulative rollout error.

Specifically, let a candidate schedule $\mathcal{S}=\{s_i\}_{i=0}^{L-1}$ define reuse spans $\mathcal{E}(\mathcal{S})=\{(s_i,s_{i+1})\}_{i=0}^{L-2}$, and let $\mathcal{C}(i,i+1)=\{t:s_i<t<s_{i+1}\}$ be the cached intermediate steps in span $(s_i,s_{i+1})$.
For calibration prompt $p$, chunk $c$, and step $t\in\mathcal{C}(i,i+1)$, the relative error between the cached prediction $\tilde{v}^{(i,i+1)}_{p,c,t}$ and the full-model prediction $v^{\mathrm{ref}}_{p,c,t}$ is:
\begin{equation}
  e_{p,c,t}(i,i+1)=
  \frac{
  \left\|
  \tilde{v}^{(i,i+1)}_{p,c,t}
  -
  v^{\mathrm{ref}}_{p,c,t}
  \right\|_2
  }{   
  \left\|v^{\mathrm{ref}}_{p,c,t}\right\|_2
  }.
\end{equation}
Since different chunks may exhibit substantially different sensitivities to the same reuse span, RCS uses a quantile statistic to obtain a robust chunk-aggregated error:
\begin{equation}
\ell_p(i,i+1,t)
=
Q_q
\left(
\{e_{p,c,t}(i,i+1) : \forall c\}
\right),
\end{equation}
where $Q_q(\cdot)$ denotes the $q$-th quantile. Then, RCS accumulates these robust step errors over the span and over all spans:
\begin{equation}
r_p(i,i+1)=\sum_{t\in\mathcal{C}(i,i+1)}\ell_p(i,i+1,t),
\qquad
R_p(\mathcal{S})=\sum_{(i,i+1)\in\mathcal{E}(\mathcal{S})} r_p(i,i+1).
\end{equation}
For cross-prompt robustness, RCS first computes the mean $\mu(\mathcal{S})$ and standard deviation $\sigma(\mathcal{S})$ of the prompt-wise errors $\{R_p(\mathcal{S}): p \in \mathcal{D}_{\mathrm{calib}}\}$ over the calibration set $\mathcal{D}_{\mathrm{calib}}$, and then employs Conditional Value-at-Risk (CVaR) as the tail risk:
\begin{equation}
\operatorname{CVaR}_{0.8}(\mathcal{S})
=
\frac{1}{M-\lceil 0.8M\rceil+1}
\sum_{m=\lceil 0.8M\rceil}^{M}
R_{(m)}(\mathcal{S}).
\end{equation}
where $R_{(m)}$ is the $m$-th smallest prompt error. Under a fixed number of function evaluations (NFE) budget $L$, RCS minimizes:
\begin{equation}
J_D(\mathcal{S})
=
\mu(\mathcal{S})
+
\eta\sigma(\mathcal{S})
+
\rho\operatorname{CVaR}_{0.8}(\mathcal{S})
+
\lambda_q Q_{\mathrm{end}}(\mathcal{S}),
\label{eq:target}
\end{equation}
\begin{wrapfigure}{r}{0.45\textwidth}
    \centering
    \includegraphics[width=\linewidth]{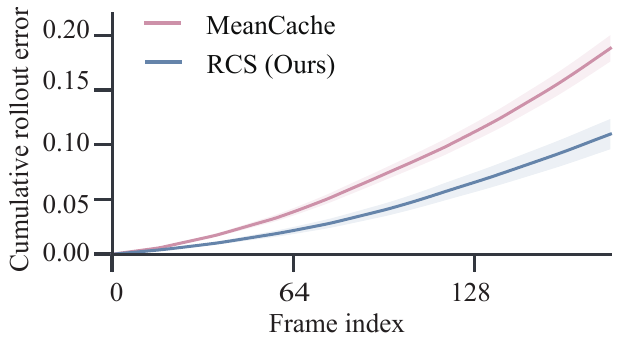}
    \vspace{-5mm}
    \caption{
    Cumulative rollout error (vertical) versus frame index (horizontal), with error accumulated from frame‑wise output RMSE. Curves compare MeanCache and our RCS, averaged over 30 prompts randomly sampled from VBench using SkyReels‑V2‑1.3B‑540P. Shaded regions indicate the standard error of the mean, and vanilla outputs serve as the reference.
    }
    \vspace{-2mm}
    \label{fig:rcs_exp}
\end{wrapfigure}
where $\eta, \rho, \lambda_q$ are trade-off hyperparameters, 
and $Q_{\mathrm{end}}(\mathcal{S})$ is an endpoint-mismatch regularizer that measures the error between the final latent produced by the cached trajectory and that of the full-model trajectory.
Here $\mu(\mathcal{S})$ encourages low average error, $\sigma(\mathcal{S})$ promotes robustness across prompts, and $\operatorname{CVaR}_{0.8}$ explicitly penalizes the average risk over the worst 0.2-fraction of prompts, thereby improving robustness to difficult inputs. The last term $Q_{\mathrm{end}}$ preserves the alignment between the final latent of the cached trajectory and that of the full-model trajectory. By solving Eq.~\ref{eq:target} using dynamic programming, we can obtain the schedule $\mathcal{S}_{\mathrm{base}}$. Details are provided in Appendix~\ref{app:dp-rcs}.

Moreover, the first chunk plays a unique role in AR generation since it is repeatedly reused as context for subsequent chunks. Consequently, its errors can affect more subsequent chunks. Thus, for the first chunk, RCS allocates an additional NFE budget $L^f$ and obtains a denser schedule $\mathcal{S}_{\mathrm{first}}$ by optimizing the same objective under this increased budget.
This design allows aggressive reuse for later chunks while providing stronger protection for the first chunk, whose approximation errors can have the largest downstream impact.
With the above designs, as shown in Fig.~\ref{fig:rcs_exp}, RCS achieves a lower cumulative rollout error than MeanCache~\citep{gao2026meancache}, whose error is higher and grows faster along the generation trajectory.

\subsubsection{Autoregressive Trajectory-Aware Guidance Scheduling}
\begin{wrapfigure}{r}{0.45\textwidth}
    \centering
    \includegraphics[width=\linewidth]{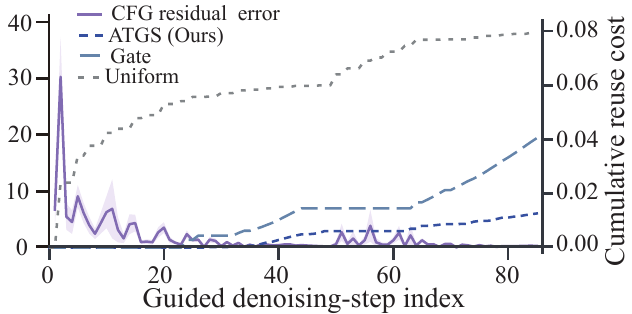}
    \vspace{-5mm}
    \caption{
    The horizontal axis shows the guided denoising step index along the global AR path. The left and right vertical axes plot the CFG residual error (purple) and cumulative reuse cost, respectively. ATGS (Ours), Gate, and Uniform are three CFG refresh schedules compared under the same budget. Results are on SkyReels‑V2‑1.3B‑540P with 30 prompts from VBench.
    }
    \vspace{-2mm}
    \label{fig:atgs_exp}
\end{wrapfigure}
Classifier-free guidance (CFG) introduces significant computational cost from repeated residual refreshes at each denoising step~\citep{ho2022classifier,lyu2025fastercache,castillo2025adaptive}.
Although existing CFG reuse strategies~\citep{lyu2025fastercache,lee2026tortoise} have proven effective for bidirectional full-attention video models, they are designed for a single denoising trajectory and assume global temporal consistency across the entire sequence. 
In AR video generation, however, each chunk follows its own denoising trajectory, and together these trajectories form a sequential path. This key difference makes independent per-chunk scheduling suboptimal, where reuse errors accumulate along the global AR path, and a decision that appears reasonable for one chunk can lead to significant cumulative error over the full path. Therefore, a globally calibrated reuse schedule covering the entire AR trajectory is essential.
%
%
%
%
%
%
%
Motivated by this observation, we propose autoregressive trajectory-aware guidance scheduling (ATGS), which formulates CFG reuse scheduling as a path-planning problem over the sequential AR denoising path. As shown in the purple lines in Fig.~\ref{fig:atgs_exp}, CFG residual error is generally small across most steps but exhibits pronounced spikes at a few critical positions. This empirical pattern suggests that residuals can be safely reused within locally smooth intervals, while refreshes must be placed where residual changes are substantial.
Consider two denoising steps $i < t$, where the CFG residual is refreshed at step $i$ and reused at a later step $t$. Let $\Delta_{p,i} \in \mathbb{R}^{D}$ and $\Delta_{p,t} \in \mathbb{R}^{D}$ denote the CFG residual for calibration prompt $p$ at step $i$ and $t$, respectively. The relative prediction error introduced by this reuse is:
\begin{equation}
\hat{\epsilon}^{\text{reuse}}_{p,i\rightarrow t} - \hat{\epsilon}^{\text{full}}_{p,t}
= (w-1)(\Delta_{p,i} - \Delta_{p,t}),
\label{eq:cfg_reuse_general}
\end{equation}
where $w$ is the fixed guidance scale. Thus, the residual error $\|\Delta_{p,i} - \Delta_{p,t}\|$ directly governs the approximation error introduced by reuse.
To quantify this error across both chunks and prompts, ATGS first defines a metric $\delta_p(i,t)$ that captures both the magnitude and directional change between steps $i$ and $t$:
\begin{equation}
\delta_p(i,t) = 
\frac{\|d_{p,t}\|_2^2 + \|d_{p,i}\|_2^2 - 2\|d_{p,t}\|_2\|d_{p,i}\|_2 \cos(d_{p,t}, d_{p,i})}{D},
\end{equation}
where $d_{p,n} = \mathrm{Flatten}(\Delta_{p,n}) \in \mathbb{R}^D$.
Thus, for each prompt $p$, the cumulative error of reusing a residual refreshed at step $i$ until the next refreshed step $j$ is $c_p(i,j) = \sum_{t=i+1}^{j-1} \delta_p(i,t)$. 
To obtain a schedule that generalizes across prompts, ATGS aggregates these errors over the calibration set $\mathcal{D}_{\mathrm{calib}}$:
\begin{equation}
c_G(i,j) = \frac{1}{|\mathcal{D}_{\mathrm{calib}}|} \sum_{p \in \mathcal{D}_{\mathrm{calib}}} c_p(i,j).
\end{equation}
Given the NFE budget $L^C$, ATGS seeks a CFG cache schedule $\mathcal{S}^C$ by selecting $L^C$ steps $\mathcal{S}^C = \{s^C_i\}_{i=0}^{L^C-1}$ that minimize the total path error:
\begin{equation}
\mathcal{S}^C = \arg\min_{\mathcal{S} \in \Omega_{L^C}} \sum_{i=0}^{L^C-1} c_G(s^C_i, s^C_{i+1}),
\label{eq:ATGS_optimization}
\end{equation}
where $\Omega_{L^C}$ denotes the set of all feasible schedules consisting of $L^C$ refresh steps along the execution path. This optimization can be solved by standard budgeted dynamic programming in $O(L^C N^2)$ time. Details are provided in Appendix~\ref{app:dp-atgs}.

During inference, the schedule $\mathcal{S}^C$ is applied deterministically. At each guided step $t$, if $t \in \mathcal{S}^C$, the CFG residual is refreshed by performing the full model forward pass. Otherwise, the residual from the most recent refresh step $i(t) = \max\{s \in \mathcal{S}^C : s \leq t\}$ is reused:
\begin{equation}
\hat{\epsilon}_t = \epsilon^c_t + (w-1) \Delta_{i(t)}.
\end{equation}
This formulation ensures that CFG refresh points are placed precisely at high-error positions along the global AR path, yielding a computationally efficient yet fidelity-preserving approximation of full CFG. As illustrated in Fig.~\ref{fig:atgs_exp}, ATGS achieves a substantially lower cumulative reuse cost than both Gate and Uniform methods~\citep{lyu2025fastercache,lee2026tortoise,castillo2025adaptive}, demonstrating its effectiveness.

\subsubsection{Spectral Structure Correction}
\begin{wrapfigure}{r}{0.43\textwidth}
    \vspace{-1mm}
    \centering
    \includegraphics[width=\linewidth]{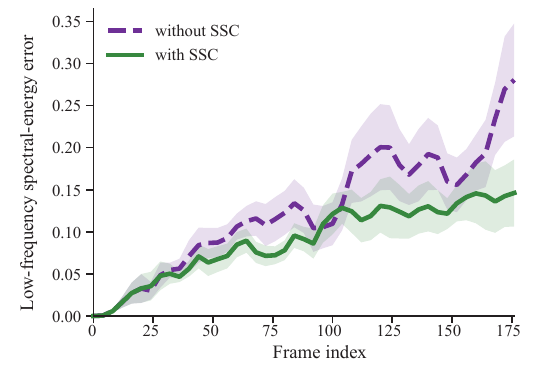}
    \vspace{-4mm}
    \caption{
    Low-frequency spectral-energy error (normalized by the first frame). Purple: cache-based rollout w/o SSC; green: same rollout with SSC enabled. SSC reduces low-frequency error growth under identical rollout settings. Results on SkyReels‑V2‑1.3B‑540P with 30 VBench prompts.}
    \label{fig:ssc_exp}
\end{wrapfigure}
Cache reuse in AR video generation accumulates low-frequency error~\citep{li2026freqforcing}, which represents global appearance and layout, and becomes the dominant error source. 
FreqForcing~\citep{li2026freqforcing} addresses this with an auxiliary branch for a low-frequency structural prior, but incurs non-negligible attention overhead and uniformly applies this prior across all regions, potentially over-constraining high-frequency areas and obscuring local motion or fine-grained details~\citep{ma2026tuning,lu2025freelong++}.

Motivated by these limitations, we propose spectral structure correction (SSC), which stabilizes low-frequency structures while preserving the motion and fine details of the current generation chunk at inference. Specifically, SSC employs the first chunk as a stable structural prior, serving as a reference for low-frequency information.
For chunk \(c\) at layer \(\ell\) and denoising step \(t\), let \(Q_{\ell,c,t}\) denote the query of the self-attention layer and \(A_{\ell,c,t}\) denote its original output. To capture only the coarse low-frequency cues while reducing computational cost, SSC introduces an auxiliary low-resolution attention path that reuses the key and value caches (\(K_{\ell,0,t}\) and \(V_{\ell,0,t}\)) from the first chunk (\(c=0\)):
\begin{equation}
A'_{\ell,c,t}
=\operatorname{Attn}\left(
\mathcal{P}(Q_{\ell,c,t}),
\mathcal{P}(K_{\ell,0,t}),
\mathcal{P}(V_{\ell,0,t})
\right),
\end{equation}
where \(\mathcal{P}\) denotes a \(4 \times 4\) pooling operator. This pooled attention captures the coarse relationship between the current query and the prior context while suppressing token-level fluctuations. Rather than replacing the original output, SSC extracts only the low-frequency discrepancy between the structural reference and the current output via a low-pass operator \(\operatorname{LP}\):
\begin{equation}
r_{\ell,c,t}
=\mathcal{U}\left(\operatorname{LP}\left(
A'_{\ell,c,t}-\mathcal{P}(A_{\ell,c,t})
\right)\right),
\end{equation}
where \(\mathcal{U}\) is the upsampling operator that upsamples the pooled representation to the original resolution. Here, \(r_{\ell,c,t}\) constitutes the coarse low-frequency structural correction suggested by the first chunk. Since high-frequency components are removed before fusion, fine-grained texture details and short-range motion remain governed by the original output \(A_{\ell,c,t}\).

To adaptively apply the low-frequency correction, SSC further introduces an adaptive correction ratio \(R_{c,t} \in [R_{\min}, 1]\). We first compute a motion score \(m_{c,t}\) by aggregating all adjacent-frame variations within the chunk:
\begin{equation}
m_{c,t}=\sum_{\forall f \in \text{chunk}\,c}\left[\frac{\left\|X_{c,t,f}-X_{c,t,f-1}\right\|_{1}}{\left\|X_{c,t,f-1}\right\|_{1}}\right],
\label{eq:ssc_motion_score}
\end{equation}
where \(X_{c,t,f}\) denotes the latent representation of the \(f\)-th frame in chunk \(c\) at step \(t\). This score serves as a robust proxy for local motion intensity. The correction ratio is then obtained by mapping this score between two fixed thresholds:
\begin{equation}
R_{c,t}=\operatorname{Clip}\left(1-(1-R_{\min})\operatorname{Clip}\left(\frac{m_{c,t}-\tau_{\mathrm{low}}}{\tau_{\mathrm{high}}-\tau_{\mathrm{low}}},0,1\right),R_{\min},1\right),
\label{eq:ssc_motion_gate}
\end{equation}
where \(\tau_{\mathrm{low}}\) and \(\tau_{\mathrm{high}}\) are the lower and upper thresholds, and \(R_{\min}\) is the minimum correction ratio. Eq.~\ref{eq:ssc_motion_gate} ensures that chunks with low motion receive a stronger injection of low-frequency information from the first chunk, whereas chunks with high motion retain only a minimal fraction of such correction. The final corrected self-attention output is thus:
\begin{equation}
\bar A_{\ell,c,t}
=A_{\ell,c,t}
+\lambda_A\operatorname{Clip}_{\rho_A}\left(r_{\ell,c,t}\odot R_{c,t}, -\rho_A, \rho_A \right),
\end{equation}
where \(\lambda_A\) is a scaling factor that controls the overall correction strength, and \(\rho_A\) is a clipping bound that prevents excessively large updates. This design enables SSC to adaptively enhance structural coherence in static or slow-moving regions while preserving the high-frequency vitality of dynamic areas. 
Fig.~\ref{fig:ssc_exp} shows that low-frequency error grows faster without SSC. Under the same cache-based setting, SSC substantially slows this accumulation.
Note that SSC operates orthogonally to the caching schedules established by RCS and ATGS. Specifically, RCS continues to dictate which denoising steps execute full model evaluations, while ATGS governs when to refresh CFG residuals.

\section{Experiments}
\label{sec:experiments}

\paragraph{Base models.}
To evaluate the efficacy of the proposed method, we select two representative autoregressive video generation backbones: SkyReels-V2-1.3B-540P and FramePack-F1~\citep{chen2025skyreels,zhang2026frame}. SkyReels-V2 adopts a hierarchical autoregressive denoising pipeline: each video section is partitioned into chunks, and earlier chunks are denoised for a predefined number of steps before subsequent chunks enter the denoising process. This execution pattern produces non-uniform denoising levels across chunks at the same sampling iteration. FramePack-F1 follows a forward-only next-frame-section prediction pipeline. It packs previous frame contexts into a fixed context budget and generates later frame sections sequentially. In our experiments, FramePack-F1 uses 25 denoising steps per generated section, 145 generated frames, and $608\times640$ resolution.

\paragraph{Baselines and Configuration.}
We compare with three training-free cache acceleration baselines: TeaCache, FlowCache, and MotionCache \citep{liu2411timestep,ma2026flow,xu2026motion}. All methods use matched prompts, seeds, video lengths, resolutions, and input conditions within each backbone. The detailed hyperparameter settings are provided in Appendix~\ref{app:implementation-details}.

\paragraph{Evaluation metrics.} For SkyReels-V2, we report VBench score on the full 946-prompt benchmark together with PSNR, SSIM, and LPIPS diagnostics. For FramePack-F1, we report the VBench++ Imaging Quality, Quality, I2V, and Total scores. Latency is measured end-to-end. We also report the speedup relative to Vanilla model inference.

\begin{table*}[t]
    \centering
    \caption{Comparison on SkyReels-V2-1.3B-540P~\citep{chen2025skyreels}.}
    \label{tab:main_results_skyreels}
    \small
    \setlength{\tabcolsep}{4pt}
    \begin{tabular}{ccccccc}
        \toprule
        Method & Latency (s)$\downarrow$ & Speedup$\uparrow$ & VBench$\uparrow$ & PSNR$\uparrow$ & SSIM$\uparrow$ & LPIPS$\downarrow$ \\
        \midrule
        Vanilla & 1608.6470 & 1.0000$\times$ & 82.9749 & -- & -- & -- \\
        TeaCache-slow & 814.0000 & 1.9762$\times$ & 81.6700 & 18.9600 & 0.6501 & \textbf{0.2472} \\
        TeaCache-fast & 686.0000 & 2.3450$\times$ & 80.0600 & 18.3900 & 0.6121 & 0.3063 \\
        FlowCache-slow & 323.4470 & 4.9734$\times$ & 82.0870 & 17.2600 & 0.6370 & 0.3145 \\
        FlowCache-fast & 268.6330 & 5.9883$\times$ & 81.9805 & 16.2437 & 0.6088 & 0.3597 \\
        MotionCache & 343.3827 & 4.6847$\times$ & 81.8000 & 17.1731 & 0.6301 & 0.3267 \\
        TRAC & \textbf{266.8501} & \textbf{6.0283$\times$} & \textbf{82.4440} & \textbf{20.1208} & \textbf{0.6882} & 0.2568 \\
        \bottomrule
    \end{tabular}
    \vspace{-4mm}
\end{table*}

\begin{table*}[t]
    \centering
    \caption{Comparison on FramePack-F1~\citep{zhang2026frame}.}
    \label{tab:main_results_framepack}
    \small
    \setlength{\tabcolsep}{4pt}
    \begin{tabular}{lcccccc}
        \toprule
        Method & Latency (s)$\downarrow$ & Speedup$\uparrow$ & Imaging Quality$\uparrow$ & Quality$\uparrow$ & I2V$\uparrow$ & Total$\uparrow$ \\
        \midrule
        Vanilla & 360.7360 & 1.0000$\times$ & 0.7055 & 0.8203 & 0.9565 & 0.8884 \\
        TeaCache-fast & 228.8670 & 1.5760$\times$ & \textbf{0.7108} & 0.8125 & 0.9557 & 0.8841 \\
        TeaCache-slow & 254.3220 & 1.4180$\times$ & 0.7048 & 0.8123 & 0.9580 & 0.8851 \\
        FlowCache-fast & 241.3940 & 1.4940$\times$ & 0.7073 & 0.8117 & \textbf{0.9609} & 0.8863 \\
        FlowCache-slow & 280.2290 & 1.2870$\times$ & 0.6804 & 0.7780 & 0.9030 & 0.8405 \\
        MotionCache & 253.9000 & 1.4210$\times$ & 0.7049 & 0.8077 & 0.9582 & 0.8830 \\
        TRAC & \textbf{216.2220} & \textbf{1.6680$\times$} & 0.7046 & \textbf{0.8205} & 0.9560 & \textbf{0.8882} \\
        \bottomrule
    \end{tabular}
    \vspace{-4mm}
\end{table*}
\subsection{Results}

\textbf{Main Results.}
As shown in Tables~\ref{tab:main_results_skyreels} and~\ref{tab:main_results_framepack}, across SkyReels-V2 and FramePack-F1, TRAC consistently provides a strong efficiency--quality trade-off. 
On SkyReels-V2, TRAC reduces dense latency from 1608.6470 seconds to 266.8501 seconds, an 83.4\% reduction corresponding to a 6.0283$\times$ speedup. Its VBench score is 82.4440, only about 0.6\% lower than Vanilla. Among compared methods, TRAC achieves the highest VBench score, a PSNR of 20.1208, and an SSIM of 0.6882. Its LPIPS of 0.2568 is second only to TeaCache-slow, while requiring substantially lower latency. Compared with FlowCache-fast at a similar latency, TRAC improves VBench score from 81.9805 to 82.4440, PSNR from 16.2437 to 20.1208, SSIM from 0.6088 to 0.6882, and LPIPS from 0.3597 to 0.2568, showing significant performance advantages.

On FramePack-F1, under the fixed FramePack configuration with $\mathrm{cfg}=1$, seed 1024, TRAC reduces the latency of dense Vanilla inference from 360.7360 to 216.2220 seconds, yielding the lowest latency and the highest speedup of 1.6680$\times$. It also achieves the highest Quality score of 0.8205. FlowCache-fast obtains the highest I2V score of 0.9609, while Vanilla obtains the highest Total score of 0.8884. TRAC reaches 0.8882, only 0.0002 lower. TeaCache-fast obtains the highest Imaging Quality score of 0.7108. Compared with TeaCache-slow, FlowCache-fast, and MotionCache, TRAC is faster and obtains higher Total scores, while maintaining a comparable I2V score. These results show that TRAC reaches the most efficient operating point while preserving aggregate quality close to dense Vanilla inference.

\begin{table}[!t]
    \centering
    \caption{Additional baseline comparison on SkyReels-V2.}
    \label{tab:baseline_screening}
    \small
    \begin{tabular}{cccccc}
        \toprule
        Method & Latency (s)$\downarrow$ & Speedup$\uparrow$ & Quality$\uparrow$ & Semantic$\uparrow$ & Total$\uparrow$ \\
        \midrule
        FreqForcing & 583.4457 & 2.7571$\times$ & \textbf{83.4603} & \textbf{82.5981} & \textbf{83.2879} \\
        FasterCache & 458.6105 & 3.5077$\times$ & 80.8739 & 77.8126 & 80.2617 \\
        MeanCache & 405.6705 & 3.9654$\times$ & 67.2080 & 61.6877 & 66.1040 \\
        TRAC & \textbf{266.8461}& \textbf{6.0284$\times$} & 82.9253 & 82.1763 & 82.7755 \\
        \bottomrule
    \end{tabular}
    \vspace{-4mm}
\end{table}

\textbf{Additional Baseline Comparison.}
To determine if the efficiency-quality behavior of TRAC holds across a broader range of acceleration mechanisms, we further evaluate it against FreqForcing, FasterCache, and MeanCache on SkyReels-V2. These selected baselines cover various complementary training-free strategies, such as spectral correction, cache reuse, and JVP-based schedule selection. Since full evaluation of every auxiliary baseline is computationally expensive, we randomly sample 94 prompts from the VBench prompt set and evaluate all methods under matched prompts, seeds, resolutions, and video lengths. Table~\ref{tab:baseline_screening} presents the comparison under a matched evaluation setting. TRAC significantly outperforms the other methods in efficiency, achieving a $6.03\times$ speedup with a latency of only 266.8s. Although FreqForcing obtains slightly higher VBench aggregates (e.g., a Total score of 83.29 vs. TRAC's 82.78), it requires a substantial 583.4s to complete. Both FasterCache and MeanCache fall behind TRAC in terms of speed and generation metrics. Consequently, TRAC achieves a superior trade-off, successfully preserving high-fidelity visual quality while providing the most aggressive speedup. In Appendix~\ref{app:additional-qualitative}, we provide quality comparisons.

\subsection{Ablation Study}
\begin{wrapfigure}{r}{0.43\textwidth}
    \centering
    \includegraphics[width=\linewidth]{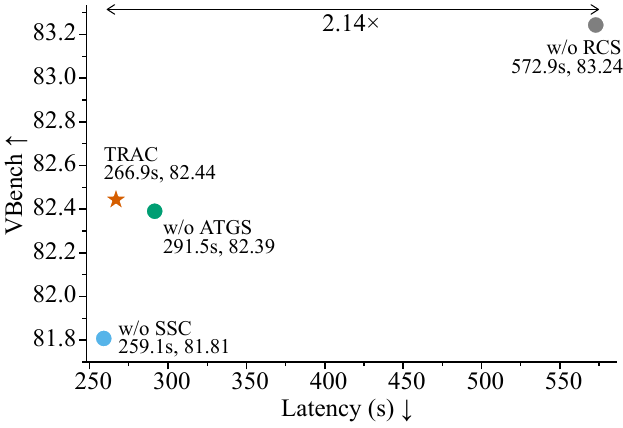}
     \vspace{-4mm}
    \caption{\textbf{Component ablation on SkyReels-V2.} The scatter plot compares latency and VBench total after removing one component at a time.}
    \label{fig:ablation_tradeoff}
    \vspace{-4mm}
\end{wrapfigure}
%
As shown in Fig.~\ref{fig:ablation_tradeoff}, we conduct the component ablation on the SkyReels-V2-1.3B-540P, using VBench score as the quality metric. The full TRAC setting combines RCS, ATGS, and SSC. Removing SSC reduces latency from 266.9 to 259.1 seconds, but lowers VBench from 82.44 to 81.81, indicating that SSC mainly contributes to quality. Removing ATGS increases latency to 291.5 seconds while maintaining a similar VBench score of 82.39, demonstrating the efficiency contribution of CFG residual reuse. Removing RCS increases latency to 572.9 seconds, more than doubling that of TRAC, despite a VBench score of 83.24. These comparisons highlight the complementary roles of RCS, ATGS, and SSC. 
Detailed sensitivity analyses of the SSC correction strength
\(\lambda_A\) and the RCS objective weights \(\eta\), \(\rho\), and \(\lambda_q\) are reported in Appendix~\ref{app:more-ablation-study}.
%

\section{Limitations and Future Work}

First, while TRAC is effective on two widely used autoregressive backbones and achieves the best trade-off between performance and speed, its generalization to emerging AR video models remains an open question. Addressing this would require considerable computational resources, which are not currently available to us. Second, although TRAC already outperforms existing caching methods in terms of speedup, it is orthogonal to complementary acceleration techniques, including quantization, distillation, and sparse attention. Integrating TRAC with these techniques may yield additional efficiency gains, a direction we leave for future work.


\section{Conclusion}

In this paper, we present trajectory-aware reuse and adaptive correction (TRAC), a training-free framework for efficient autoregressive (AR) video generation.
Existing acceleration methods mainly target single-trajectory generation with bidirectional attention. AR video generation, by contrast, sequentially couples chunk-level denoising trajectories. Consequently, approximation errors accumulate and propagate through the generation process. TRAC addresses this challenge with three components, including robust cumulative scheduling (RCS), autoregressive trajectory-aware guidance scheduling (ATGS), and spectral structure correction (SSC).
RCS selects cache reuse schedules by cumulative rollout error and cross-chunk/prompt variation. ATGS coordinates CFG refreshes along the global AR trajectory. SSC restores low-frequency structure of the first chunk to correct long-term structural loss. 
Experiments on SkyReels-V2 and FramePack-F1 show that, compared with existing methods, TRAC achieves both the highest inference efficiency and the best generation quality for AR video generation.

\subsection*{Acknowledgments}

This work was supported by the National Natural Science Foundation of China (No. 62606133).


\section*{AI Use Statement}

During the preparation of this manuscript, we used generative AI tools solely for language polishing and code-writing assistance. Generative AI tools were not used to replace experimental execution, generate experimental data, or serve as a source of unverified research conclusions or literature evidence. The authors have rigorously checked and reviewed all generated content to ensure its correctness, and they take full responsibility for the final code, content, and claims presented in this work.

\section*{Reproducibility Statement}

To facilitate reproducibility, we provide the anonymized source code in the supplementary material. The method and experimental settings are described in Sections~\ref{sec:method} and~\ref{sec:experiments}, with additional implementation details provided in Appendix~\ref{app:implementation-details}. Crucially, all proposed schedules (the shared denoising path, first-chunk schedule, and CFG refresh set) are fixed post-calibration, ensuring deterministic execution with zero online scoring overhead. All experiments use fixed random seeds and are evaluated under strictly identical conditions (e.g., prompts, resolutions, and video lengths).

\bibliography{iclr2027_conference}
\bibliographystyle{iclr2027_conference}
\clearpage
\appendix
\raggedbottom
\setlength{\intextsep}{4pt}

\section{Implementation Details}
\label{app:implementation-details}

All experiments are implemented in PyTorch and executed on NVIDIA A800 GPUs with batch size 1.

\paragraph{SkyReels-V2.}
Following the official SkyReels-V2 implementation, we use 177 frames, 2 autoregressive sections with 5 causal chunks per section, 17 overlap frames, 50 denoising steps along each chunk-local trajectory, an autoregressive step offset of 5, CFG scale $w=6.0$, shift 8.0, 24 fps, and 540P resolution. For TRAC, RCS uses $\mathcal{S}_{\mathrm{first}}$ for the first chunk and $\mathcal{S}_{\mathrm{base}}$ for subsequent chunks. The first and subsequent chunks retain 30 and 10 full-model forward passes, respectively. The RCS objective uses $q=0.90$ and $\operatorname{CVaR}_{0.8}$, with $\eta=0.03$, $\rho=0.01$, and $\lambda_q=0.003$. The reported run contains 86 conditional full-model evaluations. ATGS refreshes the CFG residual at 72 guided denoising steps and reuses the most recent cached CFG residual at the remaining 14 guided denoising steps.

\paragraph{FramePack-F1.}
Following the FramePack-F1 implementation, we use one input image, 145 generated frames, 4 generated sections, 25 denoising steps per section, a latent window size of 9, and $608\times640$ resolution. Following the fixed FramePack evaluation configuration, we use seed 1024, $\mathrm{cfg}=1$, a distilled guidance scale of $\mathrm{gs}=10$, and a CFG rescale of $\mathrm{rs}=0$. For TRAC, RCS retains 24 full-model forward passes in the first generated section and 10 full-model forward passes in each subsequent section. Across the four generated sections, this gives 54 full-model forward passes and 46 reuse positions. The RCS objective uses $q=0.90$ and $\operatorname{CVaR}_{0.8}$, with $\eta=0.03$, $\rho=0.01$, and $\lambda_q=0.003$. Since $\mathrm{cfg}=1$, ATGS is inactive during FramePack-F1 generation, and SSC is enabled during final generation. The remaining calibration and execution settings are provided in Appendix~\ref{app:implementation-details}.

\paragraph{SSC configuration.}
For SkyReels-V2, SSC uses the first chunk as the structural reference, whereas for FramePack-F1 it uses the first generated section. Both backbones use the $4\times4$ pooling operator $\mathcal{P}$, $\lambda_A=0.40$, $\rho_A=0.25$, $\tau_{\mathrm{low}}=0.08$, $\tau_{\mathrm{high}}=0.20$, and $R_{\min}=0.25$. The remaining implementation-specific SSC settings, including the attention-layer range, anchor-chunk configuration, smoothing kernel, and $\gamma$, are provided in Appendix~\ref{app:implementation-details}.

\paragraph{FramePack-F1 RCS calibration.}
For FramePack-F1, RCS calibration uses an outer JVP-based scheduler, with state-based JVP prediction at reuse positions. The observer uses a three-position warm-up and evaluates candidate jumps of up to eight positions. Schedules are selected according to the relative prediction error in the $\ell_2$ norm. The selected first-section and later-section schedules are:
\[
\mathcal{S}_{\mathrm{first}} =
\{0,1,2,3,4,5,6,7,8,9,10,11,12,13,14,15,16,17,18,19,20,21,22,24\},
\]
\[
\mathcal{S}_{\mathrm{base}} =
\{0,1,2,3,4,6,8,9,11,13\}.
\]
The first-section schedule uses the forced prefix \texttt{[0,1,2,3]}. At RCS reuse positions, the transformer is not executed, and the SSC correction branch is not invoked.

\paragraph{SSC implementation.}
The SSC correction branch is applied to attention layers 22--31. The low-resolution correction path uses one anchor chunk, with correction starting from chunk 1, and applies a $3\times3$ smoothing kernel. The motion-dependent correction ratio uses $\gamma=1.0$. The correction is applied on the conditional prediction branch and does not introduce an additional unconditional model evaluation.

\section{Dynamic Programming Details for Budgeted Schedule Search}
\label{app:budgeted-dp}

This appendix details the dynamic programming (DP) algorithms utilized to solve the budgeted schedule search problems formulated in Sections~3.2 and 3.3. 

\subsection{RCS Dynamic Programming}
\label{app:dp-rcs}

To determine the optimal retained denoising steps under a given budget, RCS precomputes the schedule score for all feasible reuse spans across the calibration set. After calibration, each feasible span is assigned an aggregated scalar score, denoted as $a_D(i,j)$, which is computed from the RCS span statistics described in Section~3.2. 

\paragraph{State Definition.} 
Let $D_D[b,j]$ denote the minimum accumulated RCS score of a partial schedule that selects exactly $b$ retained steps and terminates at step $j$. An auxiliary table $\operatorname{Prev}_D[b,j]$ records the optimal predecessor to facilitate trajectory backtracking.

\paragraph{Initialization.} 
In practice, the DP can be initialized from the fixed prefix used by the implementation. Assuming this warm-up phase ends at step $w$, we initialize the DP at this prefix boundary:
\begin{equation}
D_D[w+1, w] = 0,
\end{equation}
with all other states initialized to $+\infty$.

\paragraph{State Transition.} 
For each valid candidate step $j$ and its predecessor $i < j$, the state transition is formulated as:
\begin{equation}
D_D[b,j] = \min_{i<j}\left(D_D[b-1,i] + a_D(i,j)\right).
\label{eq:app-rcs-dp}
\end{equation}
The index $i$ that minimizes Eq.~\ref{eq:app-rcs-dp} is stored in $\operatorname{Prev}_D[b,j]$.

\paragraph{Termination.} 
Let $T_{\mathrm{end}}$ denote the terminal node of the schedule. Upon evaluating all valid paths that reach $T_{\mathrm{end}}$ under the NFE budget $B_D$, the solver selects the state $D_D[B_D, T_{\mathrm{end}}]$ yielding the minimum accumulated score. The baseline schedule $\mathcal{S}_{\mathrm{base}}$ is reconstructed by backtracking through $\operatorname{Prev}_D$. The denser first-chunk schedule $\mathcal{S}_{\mathrm{first}}$ is computed using the identical solver formulation but with the expanded budget $B_f$.

\begin{algorithm}[H]
\caption{Budgeted DP for RCS}
\label{alg:rcs-dp}
\begin{algorithmic}[1]
\REQUIRE Aggregated RCS span scores $a_D(i,j)$, NFE budget $B_D$, warm-up end step $w$, terminal node $T_{\mathrm{end}}$
\ENSURE Optimal cache schedule $\mathcal{S}_{\mathrm{base}}$
\STATE Initialize $D_D[b,j] \leftarrow +\infty$ and $\operatorname{Prev}_D[b,j] \leftarrow -1$ for all valid $b, j$.
\STATE $D_D[w+1, w] \leftarrow 0$ \COMMENT{Initialize from the fixed warm-up prefix}
\FOR{$b = w+2$ \TO $B_D$}
    \FOR{each candidate retained step $j \le T_{\mathrm{end}}$}
        \FOR{each valid predecessor $i < j$}
            \IF{$D_D[b-1,i] + a_D(i,j) < D_D[b,j]$}
                \STATE $D_D[b,j] \leftarrow D_D[b-1,i] + a_D(i,j)$
                \STATE $\operatorname{Prev}_D[b,j] \leftarrow i$
            \ENDIF
        \ENDFOR
    \ENDFOR
\ENDFOR
\STATE $\mathcal{S}_{\mathrm{base}} \leftarrow \operatorname{Backtrack}(D_D, \operatorname{Prev}_D, B_D, T_{\mathrm{end}})$
\STATE \textbf{return} $\mathcal{S}_{\mathrm{base}}$
\end{algorithmic}
\end{algorithm}

\subsection{ATGS Dynamic Programming}
\label{app:dp-atgs}

For ATGS, the optimal schedule minimizes the total CFG residual-reuse cost. The solver frames this as a fixed-budget path-planning problem. Let the ordered CFG events be $0, \dots, N-1$, and use $N$ as a terminal node. The calibrated edge score $c_G(i,j)$ evaluates the cumulative reuse error between two consecutive refresh steps $i$ and $j$. Optional mandatory nodes, such as boundary events, can be enforced by masking invalid transitions.

\paragraph{State Definition.} 
Let $D_C[b,j]$ denote the minimum accumulated CFG reuse cost of a path that ends at refresh step $j$ utilizing exactly $b$ refresh allocations. Similar to RCS, $\operatorname{Prev}_C[b,j]$ stores the previous refresh step for path reconstruction.

\paragraph{Initialization.} 
The sequence originates at the initial event $0$, which functions as the first refresh node:
\begin{equation}
D_C[1, 0] = 0,
\end{equation}
with all other states initialized to $+\infty$.

\paragraph{State Transition.} 
If the next refresh step is assigned to $j$, the CFG residual from the preceding refresh step $i$ is reused across all intermediate steps. The transition is identical in structure to RCS:
\begin{equation}
D_C[b,j] = \min_{i<j}\left(D_C[b-1,i] + c_G(i,j)\right).
\label{eq:app-atgs-dp}
\end{equation}

\paragraph{Termination.} 
After exhausting the CFG budget $L^C$, the solver identifies the valid state covering the terminal node $N$ with the minimum total cost. The final ATGS schedule $\mathcal{S}^C$ is retrieved via backtracking through $\operatorname{Prev}_C$.

\begin{algorithm}[H]
\caption{Budgeted DP for ATGS}
\label{alg:atgs-dp}
\begin{algorithmic}[1]
\REQUIRE Calibrated CFG reuse costs $c_G(i,j)$, CFG refresh budget $L^C$, terminal node $N$
\ENSURE Optimal CFG refresh schedule $\mathcal{S}^C$
\STATE Initialize $D_C[b,j] \leftarrow +\infty$ and $\operatorname{Prev}_C[b,j] \leftarrow -1$ for all valid $b, j$.
\STATE $D_C[1, 0] \leftarrow 0$ \COMMENT{Initialize the first refresh event}
\FOR{$b = 2$ \TO $L^C$}
    \FOR{each candidate refresh step $j \le N$}
        \FOR{each valid predecessor $i < j$}
            \IF{span $(i,j)$ satisfies constraints and $D_C[b-1,i] + c_G(i,j) < D_C[b,j]$}
                \STATE $D_C[b,j] \leftarrow D_C[b-1,i] + c_G(i,j)$
                \STATE $\operatorname{Prev}_C[b,j] \leftarrow i$
            \ENDIF
        \ENDFOR
    \ENDFOR
\ENDFOR
\STATE $\mathcal{S}^C \leftarrow \operatorname{Backtrack}(D_C, \operatorname{Prev}_C, L^C, N)$
\STATE \textbf{return} $\mathcal{S}^C$
\end{algorithmic}
\end{algorithm}

\subsection{Unified algorithm For TRAC}
\label{app:Unified algorithm}

Algorithm~\ref{alg:calibration} summarizes the offline calibration stage, which produces the RCS schedules $\mathcal{S}_{\mathrm{base}}$ and $\mathcal{S}_{\mathrm{first}}$, ATGS' $\mathcal{S}^C$, and the SSC low-frequency correction branch. Algorithm~\ref{alg:inference} describes the inference stage that applies RCS, ATGS, and SSC during chunk-wise video generation.

\begin{algorithm}[H]
\caption{Offline calibration}
\label{alg:calibration}
\begin{algorithmic}[1]
\REQUIRE calibration prompts $\mathcal{D}_{\mathrm{calib}}$, autoregressive generator $F_\theta$, NFE budgets $L, L^f, L^C$
\ENSURE cache schedules $\mathcal{S}_{\mathrm{base}}$, $\mathcal{S}_{\mathrm{first}}$, CFG cache schedule $\mathcal{S}^C$, fixed SSC correction branch
\STATE Collect the full-model predictions $v^{\mathrm{ref}}_{p,c,t}$ and the cached predictions $\tilde{v}^{(i,i+1)}_{p,c,t}$ on chunk-local denoising trajectories.
\STATE Compute the RCS span errors $r_p(i,j)$ and solve RCS under the NFE budget $L$ to obtain $\mathcal{S}_{\mathrm{base}}$.
\STATE Solve RCS under the larger first-chunk NFE budget $L^f$ to obtain the denser schedule $\mathcal{S}_{\mathrm{first}}$.
\STATE Fix the execution trace induced by $\mathcal{S}_{\mathrm{first}}$ and $\mathcal{S}_{\mathrm{base}}$, and collect the full CFG residuals $\Delta_{p,t}$ on this trace.
\STATE Aggregate the residual-reuse costs $c_G(i,j)$ over the calibration set $\mathcal{D}_{\mathrm{calib}}$ and solve the budgeted ATGS optimization under the NFE budget $L^C$ to obtain $\mathcal{S}^C$.
\STATE Fix the SSC low-frequency correction branch.
\end{algorithmic}
\end{algorithm}

\begin{algorithm}[H]
\caption{Inference}
\label{alg:inference}
\begin{algorithmic}[1]
\REQUIRE calibrated cache schedules $\mathcal{S}_{\mathrm{first}}$, $\mathcal{S}_{\mathrm{base}}$, CFG cache schedule $\mathcal{S}^C$, fixed SSC correction branch, CFG scale $w$
\ENSURE generated latent chunks $\{X_{c,T-1}\}_{c=0}^{C-1}$
\FOR{each generated chunk $c$}
    \STATE Use $\mathcal{S}_{\mathrm{first}}$ for the first chunk and $\mathcal{S}_{\mathrm{base}}$ thereafter.
    \FOR{each guided denoising step $t$}
        \IF{the current step is an RCS full-evaluation step}
            \STATE Compute the conditional prediction using the SSC path when enabled.
        \ELSE
            \STATE Reuse the cached or JVP-predicted conditional output and skip SSC.
        \ENDIF
        \IF{the step is in $\mathcal{S}^C$ or has no compatible cached residual}
            \STATE Compute the unconditional prediction $\epsilon^{uncond}_{c,t}$ and refresh the CFG residual $\Delta_{c,t}$.
        \ELSE
            \STATE Reuse the latest compatible CFG residual $\Delta_{c,i(t)}$.
        \ENDIF
        \STATE Combine the conditional prediction with the refreshed or corrected residual: $\hat{\epsilon}_{c,t}=\epsilon^{cond}_{c,t}+(w-1)\Delta_{c,i(t)}$.
    \ENDFOR
\ENDFOR
\end{algorithmic}
\end{algorithm}

\section{Qualitative Results}
\label{app:additional-qualitative}

Here, following prior cache-acceleration studies~\citep{ma2026flow,xu2026motion}, we provide matched-prompt frame comparisons. All methods use matched prompts and seeds.
Specifically, we provide eight matched-prompt qualitative comparisons on SkyReels-V2. Each panel shows four sampled frames for Vanilla, TeaCache-slow, FlowCache-slow, MotionCache, and TRAC under the same prompt and seed.

\begin{figure}[H]
    \centering
    \includegraphics[width=\textwidth]{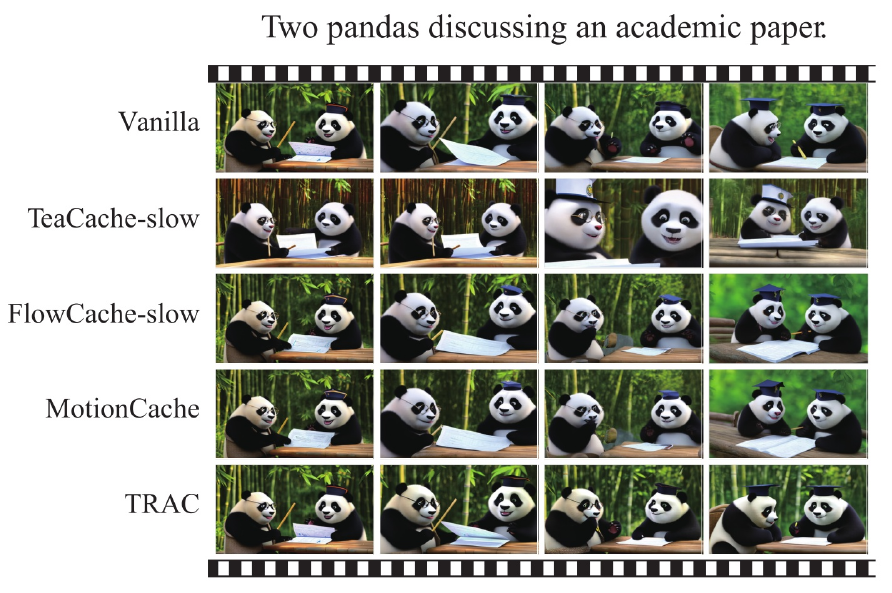}
    \caption{\textbf{Qualitative comparison on SkyReels-V2.} Prompt: ``Two pandas discussing an academic paper.''}
    \label{fig:qualitative_app_pandas_paper}
\end{figure}

\begin{figure}[H]
    \centering
    \includegraphics[width=\textwidth]{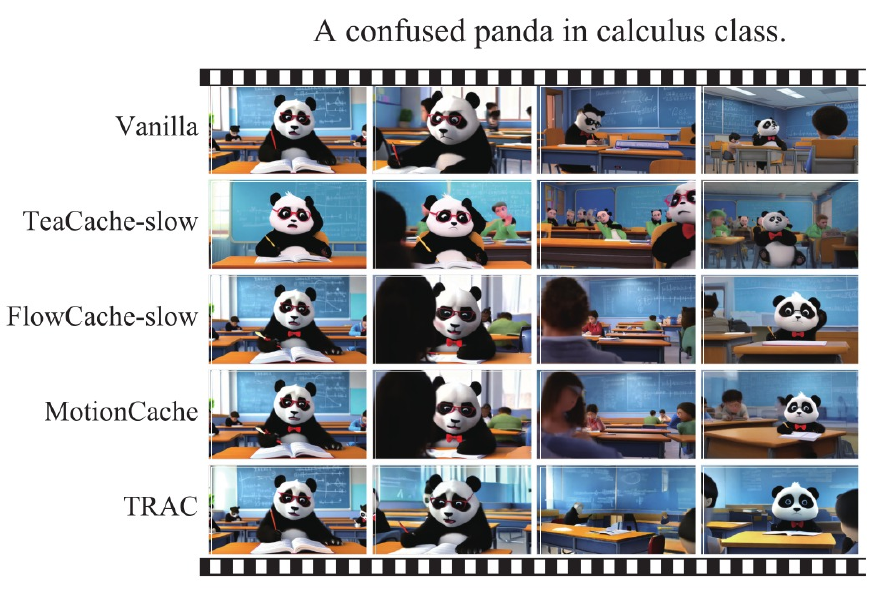}
    \caption{\textbf{Qualitative comparison on SkyReels-V2.} Prompt: ``A confused panda in calculus class.''}
    \label{fig:qualitative_app_panda_calculus}
\end{figure}

\begin{figure}[H]
    \centering
    \includegraphics[width=\textwidth]{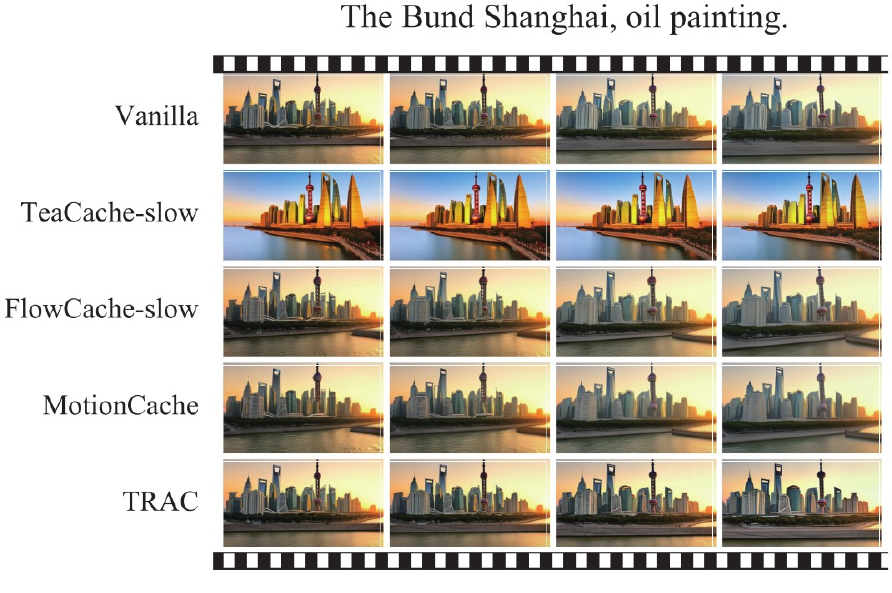}
    \caption{\textbf{Qualitative comparison on SkyReels-V2.} Prompt: ``The Bund Shanghai, oil painting.''}
    \label{fig:qualitative_app_bund_shanghai}
\end{figure}

\begin{figure}[H]
    \centering
    \includegraphics[width=\textwidth]{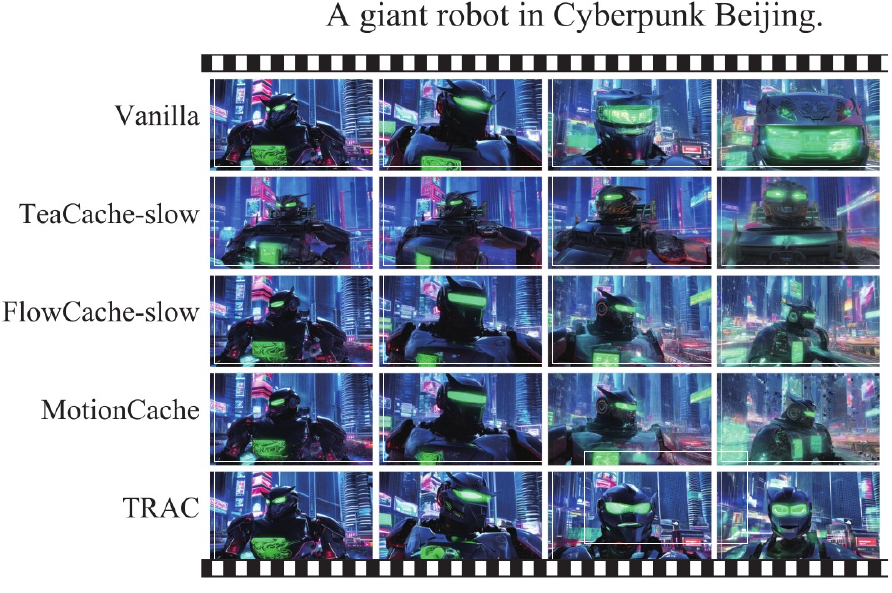}
    \caption{\textbf{Qualitative comparison on SkyReels-V2.} Prompt: ``A giant robot in Cyberpunk Beijing.''}
    \label{fig:qualitative_app_cyberpunk_robot}
\end{figure}

\begin{figure}[H]
    \centering
    \includegraphics[width=\textwidth]{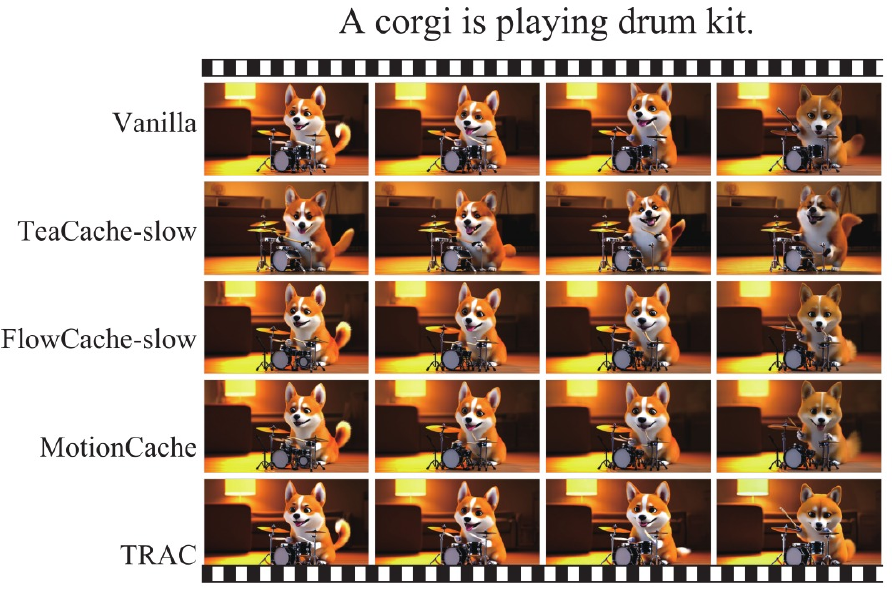}
    \caption{\textbf{Qualitative comparison on SkyReels-V2.} Prompt: ``A corgi is playing drum kit.''}
    \label{fig:qualitative_app_corgi}
\end{figure}

\begin{figure}[H]
    \centering
    \includegraphics[width=\textwidth]{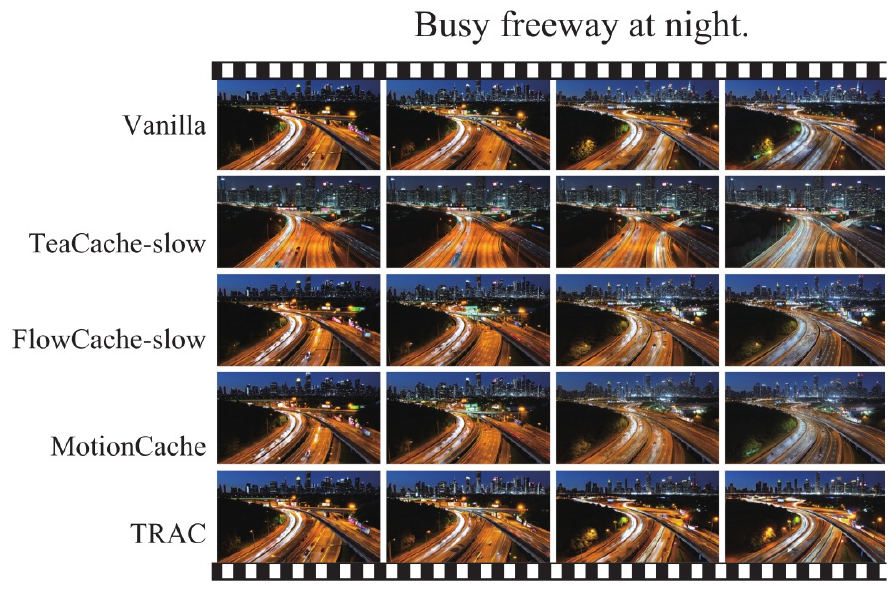}
    \caption{\textbf{Qualitative comparison on SkyReels-V2.} Prompt: ``Busy freeway at night.''}
    \label{fig:qualitative_app_freeway}
\end{figure}

\begin{figure}[H]
    \centering
    \includegraphics[width=\textwidth]{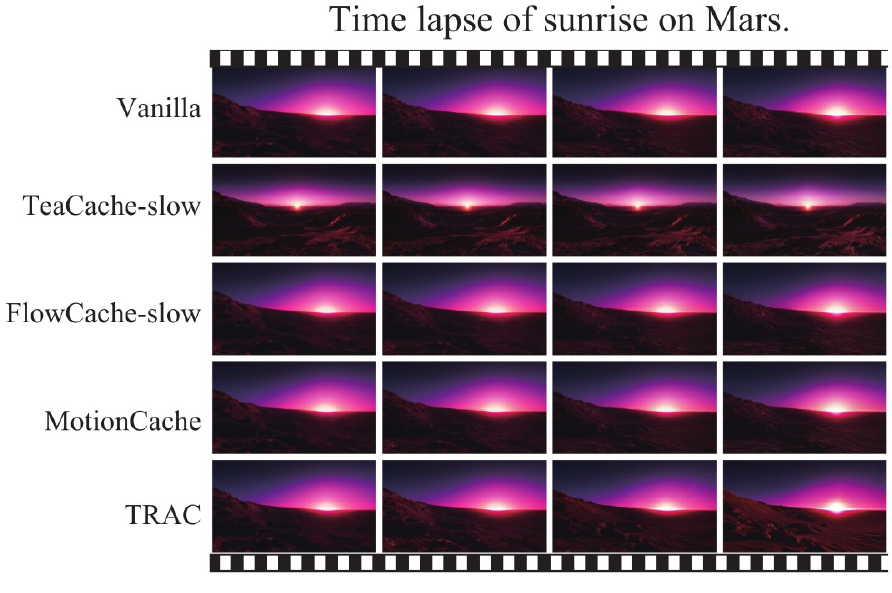}
    \caption{\textbf{Qualitative comparison on SkyReels-V2.} Prompt: ``Time lapse of sunrise on Mars.''}
    \label{fig:qualitative_app_sunrise_mars}
\end{figure}

\begin{figure}[H]
    \centering
    \includegraphics[width=\textwidth]{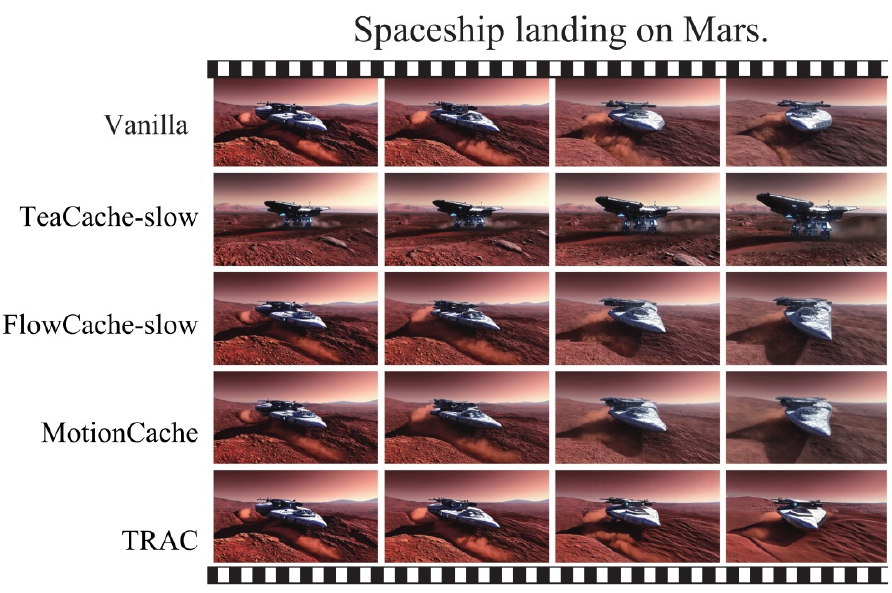}
    \caption{\textbf{Qualitative comparison on SkyReels-V2.} Prompt: ``Spaceship landing on Mars.''}
    \label{fig:qualitative_app_spaceship_mars}
\end{figure}

\section{Additional Ablation Studies}
\label{app:more-ablation-study}

We evaluate the sensitivity of the SSC correction strength and the RCS objective weights on the matched 94-prompt SkyReels-V2 subset. Unless otherwise stated, the remaining settings follow the default configuration. In Table~\ref{tab:ssc_lambda}, $\lambda_A=0.40$ gives the highest Total score and is used in the reported experiments.
\begin{table}[H]
    \centering
    \caption{Ablation studies on $\lambda_A$.}
    \label{tab:ssc_lambda}
    \small
    \begin{tabular}{lccc}
        \toprule
        $\lambda_A$ & Quality$\uparrow$ & Semantic$\uparrow$ & Total$\uparrow$ \\
        \midrule
        0.20 & 82.4367 & 81.7599 & 82.3013 \\
        0.30 & 82.7358 & 81.7433 & 82.5373 \\
        \textbf{0.40} & \textbf{82.9253} & 82.1763 & \textbf{82.7755} \\
        0.50 & 82.0764 & 82.4972 & 82.1606 \\
        0.60 & 82.0731 & \textbf{83.4813} & 82.3547 \\
        \bottomrule
    \end{tabular}
\end{table}

We next vary the three RCS objective weights independently. The other two weights are fixed to their default values in each table. Some sweep settings select the same calibrated schedule after the discrete path search, and therefore lead to identical evaluation scores.
\begin{table}[H]
    \centering
    \caption{Sensitivity to the RCS weight $\eta$, with $\rho=0.01$ and
    $\lambda_q=0.003$ fixed.}
    \label{tab:reward_eta_sweep}
    \small
    \begin{tabular}{lccc}
        \toprule
        $\eta$ & Quality$\uparrow$ & Semantic$\uparrow$ & Total$\uparrow$ \\
        \midrule
        0.03 & 82.9253 & \textbf{82.1763} & \textbf{82.7755} \\
        3.00 & \textbf{82.9524} & 81.9728 & 82.7565 \\
        10.00 & 82.9309 & 81.9741 & 82.7395 \\
        \bottomrule
    \end{tabular}
\end{table}

\begin{table}[H]
    \centering
    \caption{Sensitivity to the RCS weight $\rho$, with $\eta=0.03$ and
    $\lambda_q=0.003$ fixed.}
    \label{tab:reward_rho_sweep}
    \small
    \begin{tabular}{lccc}
        \toprule
        $\rho$ & Quality$\uparrow$ & Semantic$\uparrow$ & Total$\uparrow$ \\
        \midrule
        0.01 & 82.9253 & \textbf{82.1763} & \textbf{82.7755} \\
        3.00 & \textbf{82.9524} & 81.9728 & 82.7565 \\
        20.00 & \textbf{82.9524} & 81.9728 & 82.7565 \\
        \bottomrule
    \end{tabular}
\end{table}

\begin{table}[H]
    \centering
    \caption{Sensitivity to the RCS weight $\lambda_q$, with $\eta=0.03$
    and $\rho=0.01$ fixed.}
    \label{tab:reward_lambda_q_sweep}
    \small
    \begin{tabular}{lccc}
        \toprule
        $\lambda_q$ & Quality$\uparrow$ & Semantic$\uparrow$ & Total$\uparrow$ \\
        \midrule
        0.003 & \textbf{82.9253} & 82.1763 & \textbf{82.7755} \\
        0.30 & 82.6695 & \textbf{82.7072} & 82.6770 \\
        1.00 & 82.6695 & \textbf{82.7072} & 82.6770 \\
        \bottomrule
    \end{tabular}
\end{table}

\end{document}